\documentclass[10pt,journal,compsoc]{IEEEtran}
\ifCLASSOPTIONcompsoc
  \usepackage[nocompress]{cite}
\else
  \usepackage{cite}
\fi
\ifCLASSINFOpdf
  \usepackage[pdftex]{graphicx}
\else
\fi
\usepackage{amsmath}
\usepackage{array}
\usepackage{amsfonts}
\usepackage[linesnumbered,ruled,vlined]{algorithm2e}
\usepackage{xcolor}
\usepackage{enumerate}
\usepackage{multirow}
\usepackage{subcaption}
\usepackage[switch]{lineno}
\usepackage{tabularx}
\usepackage{enumitem}

\newcommand\clearrow{\global\let\rowmac\relax}
\clearrow
\newcolumntype{P}[1]{>{\centering\arraybackslash}p{#1}}
\DeclareMathOperator*{\argmax}{argmax} 

\newcommand{\Updated}[1]{\textcolor{black}{#1}}
\newcommand{\methodname}{\textcolor{black}{DeepAltTrip}}

\makeatletter
\newcommand{\removelatexerror}{\let\@latex@error\@gobble}
\makeatother

\begin{document}
%
\title{\methodname: Top-k Alternative Itineraries for Trip Recommendation}
%
%
%
%

\author{Syed~Md.~Mukit~Rashid,
        Mohammed~Eunus~Ali,
        and~Muhammad~Aamir~Cheema
\IEEEcompsocitemizethanks{\IEEEcompsocthanksitem S.M.M.Rashid and M.E.Ali are with Bangladesh University of Engineering And Technology, Dhaka. Emails: mukitrashid270596@gmail.com; eunus@cse.buet.ac.bd\protect\\
\IEEEcompsocthanksitem M.A.Cheema is with Monash University, Australia. E-mail: aamir.cheema@monash.edu}
}

%
%

\markboth{(PREPRINT) IEEE TRANSACTIONS ON KNOWLEDGE AND DATA ENGINEERING, UNDER REVIEW, SEPTEMBER 2021}%
{Shell \MakeLowercase{\textit{et al.}}: Bare Demo of IEEEtran.cls for Computer Society Journals}
%



\IEEEtitleabstractindextext{%
\begin{abstract}
Trip itinerary recommendation finds an ordered sequence of Points-of-Interest (POIs) from a large number of candidate POIs in a city. In this paper, we propose a deep learning-based framework, called \methodname, that learns to recommend top-$k$ alternative itineraries for given source and destination POIs. These alternative itineraries would be not only popular given the historical routes adopted by past users but also dissimilar (or diverse) to each other. The \methodname~ consists of two major components: (i)  \textit{Itinerary Net} (ITRNet) which estimates the likelihood of POIs on an itinerary by using graph autoencoders and two (forward and backward) LSTMs; and (ii) a route generation procedure to generate $k$ diverse itineraries passing through relevant POIs obtained using ITRNet. For the route generation step, we propose a novel sampling algorithm that can seamlessly handle a wide variety of user-defined constraints. To the best of our knowledge, this is the first work that \emph{learns} from historical trips to provide a set of alternative itineraries to the users. Extensive experiments conducted on eight popular real-world datasets show the effectiveness and efficacy of our approach over state-of-the-art methods.

\end{abstract}

}

\maketitle

\IEEEdisplaynontitleabstractindextext

%
\IEEEpeerreviewmaketitle

\IEEEraisesectionheading{\section{Introduction}\label{sec:introduction}}

\IEEEPARstart{D}{eciding} suitable itineraries is often challenging for tourists in an unfamiliar city. Due to the popularity of location-based social networks, an unprecedented volume of historical trips and itineraries,  each represented as an ordered sequence of Points-Of-Interest (POIs) visited, has become available. This opens up a new avenue to learn popular and suitable itineraries from the historical trip\footnote{Trips, routes, and itineraries can be used interchangeably; however, hereafter we use trips/routes to denote historical paths of users and itineraries to refer to the recommended paths.} data. In the past few years, many techniques~\cite{lim2015personalized,chen2016learning,gao2019deeptrip,wang2019empowering} have been proposed that learn from historical trips in the city and recommend the most popular itinerary from a given source $s$ to a given destination $d$. Intuitively, such a popular itinerary system returns a sequence of POIs which has been most frequently adopted by past users while traveling from $s$ to $d$. 

\Updated{
However, recommending a single itinerary is often too restrictive and may not meet a user's needs. Therefore, it is preferable to recommend multiple alternative itineraries. Quality alternative itineraries must not only be popular, but also dissimilar (or diverse) to each other. Without loss of generality,  we use  the terms diversity and dissimilarity interchangeably as both refer to the alternate itineraries with minimum overlap. 
In this paper, we propose learning-based techniques to report $k$ alternative itineraries that are popular and are also dissimilar to each other. To the best of our knowledge, we are the first to \emph{learn} multiple quality alternative itineraries from historical routes that are both popular and diverse with each other at the same time.
}
\begin{figure}[ht]
\begin{center}
\includegraphics[width=\columnwidth,height=0.75\columnwidth]{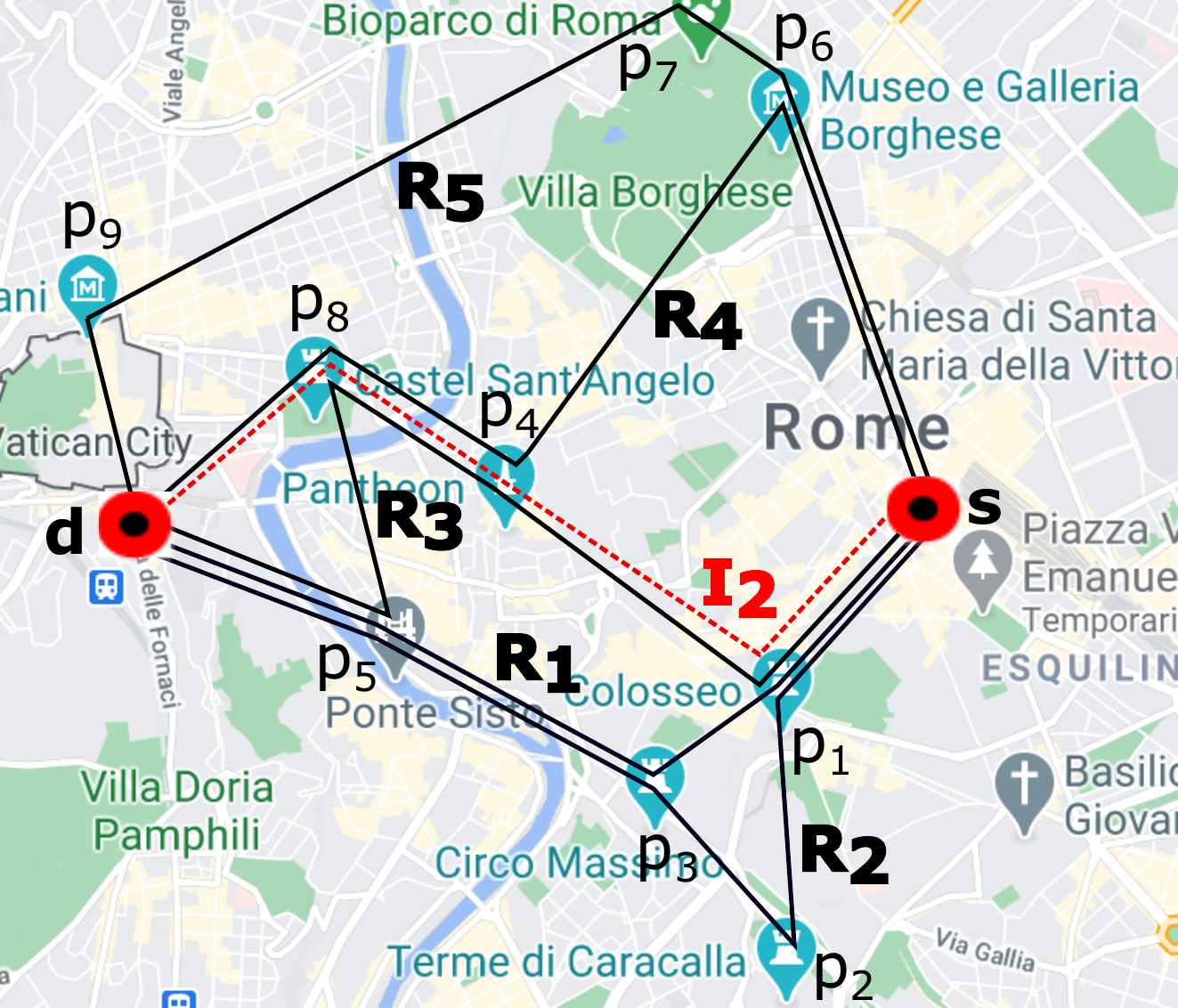}
\caption{\label{intro_exp} Different routes in Rome, each from the Central Station $s$ to a hotel $d$, passing through POIs $p_1$ to $p_9$.}
\end{center}
\end{figure}

\textbf{Motivating Example:} Figure \ref{intro_exp} shows an example where, for a source $s$ (Rome Central Station) and a destination $d$ (a hotel), five different historical routes ($R_1$ to $R_5$) are shown in solid black colored lines that pass through nine POIs in total ($p_1$ to $p_9$). Assume that the routes in descending order of popularity are given as $R_1$, $R_2$, $R_3$, $R_4$ and $R_5$.  Existing systems that return the most popular itinerary would learn to return $R_1$. If the user wants the top-$2$ popular itineraries, $R_1$ and $R_2$ would be recommended which are both very similar to each other as $R_1=(s,p_1,p_3,p_5,d)$ and $R_2=(s,p_1,p_2,p_3,p_5,d)$. This may not be desirable for a user looking for alternative itineraries to choose from. Also, if a system attempts to return diverse itineraries not considering their popularity, it may return $R_1$ and $R_5=(s,p_6,p_7,p_9,d)$. This may also be not desirable, as itinerary $R_5$ is not well supported by the historical trips. In this case, a better solution is to return two popular yet diverse itineraries such as $R_1=$ $(s,p_1,p_2,p_3,p_5,d)$ and $I_2=$ $(s,p_1,p_4,p_8,d)$. These two will be considered as quality alternatives by the querying user. Note that $I_2$ (shown in red dotted lines) does not exist in the historical routes. Our learning-based algorithms are able to discover popular and diverse itineraries that do not necessarily exist in the historical trips (e.g., $I_2$).

\textbf{Limitations of Existing Works:} All existing \emph{learning-based} itinerary recommendation systems \cite{lim2015personalized,chen2016learning,gao2019deeptrip,wang2019empowering} are designed to recommend the most popular itinerary and, to the best of our knowledge, there does not exist any learning-based technique to recommend multiple alternative itineraries.

There exists some  \emph{search-based} techniques  \cite{liang2018top,xu2019diversifying,chondrogiannis2018finding,liu2017finding} that return a set of diverse itineraries based on pre-defined popularity and/or diversity objective functions. However, these techniques are not applicable to the problem studied in this paper because the notion of diversity used in these techniques is different from ours. For example, \cite{liang2018top} aims to return routes such that the POIs within a route have diverse features. In contrast, we consider two itineraries to be more diverse if they have a smaller overlap (i.e., have fewer common POIs), so that they would be considered as alternatives with respect to each other. Also, \cite{xu2019diversifying} defines diversity to be the minimum Euclidean distance of any two POIs in two different itineraries. Thus, two itineraries that have even one common POI are considered to have zero diversity even if all other POIs are very different and far from each other.

\Updated{
Furthermore, these search-based techniques suffer from a number of limitations. First of all, most of these works only consider optimising either the popularity or the diversity, and do not take both into account at the same time while recommending a set of itineraries. However as we explain in our motivating example, attempting to optimize only diversity or only popularity without considering the other would not lead to quality alternative itineraries. Secondly, as noted in~\cite{li2021comparing}, it is not trivial to define quantitative measures to evaluate the quality of alternative itineraries and there is no agreed definition of what constitutes a set of high-quality alternative itineraries. But these search-based techniques typically require explicit modeling of popularity and/or diversity. Users may not have any prior knowledge to define and tune such metrics and, more importantly, these techniques may not be able to recommend suitable itineraries if the user fails to do so. Thirdly, since these systems do not \emph{learn} from the historical trips, they cannot incorporate the semantics of the sequence of visits in their solution. Last but not the least, these algorithms are unable to handle  user-defined constraints which limit their applicability. 
}


\textbf{Our Contributions: }
We address the above limitations and propose two novel deep-learning-based algorithms, called \textit{\methodname-LSTM} and \textit{\methodname-Samp},  that \emph{learn} from the historical trips and recommend $k$ alternative itineraries that are both popular and diverse. A key component of these algorithms is the \textit{Itinerary Net} (ITRNet), which estimates the likelihood of different POIs to be in an itinerary, using two LSTMs.


Also, both of our algorithms are \emph{metric-agnostic} in the sense that these do not require or rely on any specific diversity or popularity metrics, i.e., the users do not need to worry about defining suitable popularity or diversity metrics. Nevertheless, we extensively evaluate the algorithms using some widely used popularity and diversity metrics on real-world datasets, and these experimental studies show that both algorithms recommend high quality alternatives.

In many real-world applications, users may want to impose certain constraints on the alternative itineraries recommended by the system. For example, the total cost to visit the POIs (including traveling cost and entry tickets etc.) in each recommended itinerary must not be more than a user defined budget or each recommended itinerary must pass through some specific ``must-see'' POIs chosen by the user etc. Existing learning-based algorithms are unable to trivially handle such constraints. We propose a novel sampling algorithm, \methodname-Samp,  that can seamlessly handle a wide variety of such user defined constraints. It employs an enhanced Markov Chain Monte Carlo (MCMC) algorithm, a variation of Gibbs sampling \cite{gibbs1984}, which facilitates in pruning the candidate itineraries that do not meet the user defined criteria.

Our contributions in this paper are summarized below.
\begin{itemize}
    \item We are the first to present learning-based algorithms called, \textit{\methodname-LSTM} and \textit{\methodname-Samp}, to recommend $k$ alternative itineraries using historical trips without requiring any explicit popularity or diversity modeling.
    \item A unique advantage of our algorithm \textit{\methodname-Samp} is that it can seamlessly support additional constraints on the generated itineraries.
    \item We conduct an extensive experimental study using 8 real world datasets drawn from two domains and evaluate the algorithms on several widely used popularity and diversity metrics. The results demonstrate that our metric-agnostic algorithms propose high quality results and significantly outperform the competitors.
\end{itemize}
\label{sub:introduction}
\section{Related Works}
To our knowledge, there is no work that directly solves the problem of recommending multiple alternative itineraries through learning from historical trips. However, there are three realms of work relevant to our problem: traditional trip recommendation system that recommends a single itinerary, either through heuristics or through learning; search based techniques that attempt to return top $k$ itineraries based on an optimization problem to maximize an explicitly defined objective function; and POI recommendation methods that are mostly focused on recommending individual POIs rather than itineraries.
\subsection{Trip Recommendation Systems} 
Earlier works model tour recommendation as an orienteering problem, where the goal is to find an itinerary that maximizes a certain objective function (e.g., popularity) satisfying given constraints (e.g., budget). \cite{de2010automatic} constructs itineraries based on user visits by first constructing a POI graph and then generating an itinerary based on this graph that maximizes the total POI popularity within the user budget. \cite{chen2011discovering} argues that popular routes cannot be properly inferred only through counting from historical routes, and proposes a heuristic solution by first obtaining a transfer network and then inferring itineraries based upon an absorbing Markov Chain model built based on the network. \cite{brilhante2013shall} recommends itineraries based on user budget, time limitations and past historical data.  \cite{chen2014tripplanner} uses a two phased approach where it first interacts with the user to know her venue specifications and then uses crowd-sourced data to generate personalized POIs for the user. \cite{lim2015personalized,lim2018personalized} provide personalized user recommendation through modeling the problem as an integer programming problem given the budget constraints.
\cite{chen2016learning} is the first to \textit{learn} POI preferences and optimize the itinerary based on historical trip data and various features such as POI category, distance, and visiting time. Several other variations of the trip recommendation problem have been studied~\cite{quercia2014shortest, lim2017personalized,lim2019tour}. A comprehensive survey on this group of works is presented in ~\cite{lim2019tour}. 
Since human mobility is correlated with the location and category of POIs,~\cite{ouyang2018non} and \cite{gao2019deeptrip} proposed an adversarial model to generate a itinerary for a user query. \cite{wang2019empowering} provide a personalized itinerary through a Nerualized A* search using LSTM and self attention to estimate an observable cost and an MLP leveraging graph attention models to estimate the heuristic cost. All of the above methods provide a single itinerary for a given source and destination, and cannot be trivially extended for providing $k$ diverse and popular itineraries.

\subsection{Search Based Techniques for $k$ Itineraries} 
This realm of work adopt search based techniques to provide $k$ itineraries. Liang et al.\cite{liang2018top} provide top $k$ itineraries through searching, where they use a sub-modular function to specify diversity requirement. 
Xu et al.~\cite{xu2019diversifying} compute itineraries maintaining a minimum spatial distance covering a set of POI categories, where  the objective is to maximize the popularity satisfying the diversity constraints. Wang et al.\cite{wang2019semantic} leverage POI semantics information to develop an efficient algorithm for providing $k$ itineraries with the least cost. \cite{wang2017answering} provides top-$k$ trajectories based on user suggested location and category keyword preference. Major focus in all these works is to reduce query processing time for search algorithms, where they gather statistical POI data and require explicit diversity constraints to guide the search process.

Another group of works attempt to determine alternative routes for shortest path queries through searching in a road network graph. \cite{chen2007reliable,cheng2019shortest} leverage penalty based techniques to increase edge weights of previously used paths to gain k shortest paths which maintain some diversity.\cite{jones2012method} creates separate shortest-path trees from the source and destination nodes. The connecting branches, called \textit{plateaus} are considered for alternative routes, as longer plateaus tend to have higher dissimilarity.\cite{chondrogiannis2018finding,liu2017finding} define a dissimilarity function and then attempt to find alternative paths that exceed a pre-defined user dissimilarity threshold. In contrast to these works, the focus of our work is to develop a learning based approach that finds popular and diverse itineraries based on the historical trips.
\subsection{POI Recommendations}
Another group of works orthogonal to our research is next POI recommendation \cite{liu2016predicting,feng2018deepmove,zhao2020go}, where the objective is to recommend one or multiple POIs to visit next based on user's preferences.

Another set of works include package-POI recommendations \cite{benouaret2016package,han2017geographical}, which provide diversity in a group of POIs in a region. However none of above techniques focus on recommendation of itineraries (i.e., an ordered sequence of POIs).\label{sub:related_works}
\section{Problem Formulation}
Let $Q=\{p_1,p_2,...,p_N\}$ be the list of $N$ POIs in a city. Each POI $p_i \in Q$ is represented as a $(loc, cat)$ pair, where $loc$ is the location of $p_i$ represented by its latitude and longitude co-ordinates $\langle lat,long\rangle$ and $cat \in C$ is the category type of $p_i$. Let $\mathcal{D}$ be a multiset containing all historical routes of the past users visiting POIs in $Q$. Each route $R \in \mathcal{D}$ is an ordered sequence $(r_1, r_2, ..., r_{T})$ of POI visits, where $r_t \in Q$ is the POI at position $t$ in the route and $T$ is the total number of POIs in the route.
 
Given historical routes $\mathcal{D}$ and a query $q(s,d,k)$ with starting location $s \in Q$, ending location $d \in Q$ and an integer $k$, our aim is to recommend a set of $k$ alternative itineraries $\{I_1, I_2, ..., I_k\}$ that are both \emph{diverse} and \emph{popular}, where each itinerary $I_i$ is an ordered sequence of POIs $(r_1, r_2, ..., r_{|I_i|})$ with $r_1=s$ and $r_{|I_i|}=d$.

Intuitively by \emph{popular} itineraries we mean that the sequence of POIs visited have been frequently adopted by past users going from $s$ to $d$, and by diversity or dissimilarity we mean that the set of recommended itineraries have minimal overlap. Specific metrics to evaluate popularity and diversity are mentioned in Section \ref{sub:eval_metrics}. 

Note that unlike in a search based procedure a recommended itinerary $I_i$ is not necessarily a historical route, i.e., it is possible that $I_i\notin \mathcal{D}$. Also note that we attempt to \emph{learn} to recommend alternative itineraries rather than attempting to maximize any popularity or diversity metric. 




\label{sub:problem_formulation}
\section{Our Approach}\label{sub:our_approach}
The \methodname~ consists of two main components: (i) the \textit{Itinerary Net} (ITRNet) to estimate the probabilities of POIs at a particular position of a given itinerary by using two (forward and backward) LSTMs, and (ii) an itinerary generation algorithm to generate $k$ alternative itineraries passing through prominent POIs obtained using the ITRNet. For itinerary generation we propose two variants of \methodname: first one is an LSTM based itinerary generation technique, and the second on is a sampling based technique that provides flexibility to accommodate user constraints.
\begin{table}[ht]
\centering
\begin{tabular}[t]{|c|P{6cm}|}
\hline
   \textbf{Notation} & \textbf{Meaning}\\
\hline
   $\mathcal{D}$ & Dataset consisting of historical routes\\
\hline
   $Q$ & List $\{p_i, p_2, ..., p_N\}$ of $N$ POIs in $\mathcal{D}$\\
\hline
   $q(s,d,k)$ & User query with source POI $s$, destination POI $d$ and $k$ no. of itineraries\\
\hline
     $S_{gr}$ & Ground truth set of routes with unique $(s,d)$ pair\\
 \hline
     $S_{REC}$ & Recommended set of itineraries for query $q$\\
\hline
   $k$ & No. of itineraries in $S_{REC}$\\
\hline
   $L$ & length of recommended itineraries\\
\hline
  $p_i$ & $i^{th}$ POI in POI list $Q$\\
\hline
   $p^{(s)},p^{(d)}$ & Input source and destination POI to the model\\
\hline
    $(r_1,\cdots,r_{T})$ & Route with length $T$ \\
\hline
   $r_{a:b}$ & Shorthand for route $(r_a,r_{a+1},\cdots, r_b)$ where $b>a$\\
\hline
    $t_s,t_d$ & Position of source and destination POI\\
 \hline
     $r_t$ & POI at position $t$\\
\hline
    $p^{(p)}$ & Prominent POI\\
\hline
    $R^{(j)}$ & Itinerary generated at iteration $j$\\
\hline
    $r_t^{(j)}$ & POI at position $t$ on the itinerary $R^{(j)}$\\
 \hline
     $R_{i}$ & $i^{th}$ route in $\mathcal{D}$ or $S_{gr}$\\
 \hline
     $O[p_i]$ & No. of occurrences of POI $p_i$\\
\hline
\end{tabular}
\caption{Table of Notations}
\label{tab:notation_table}
\end{table}

\subsection{ITRNet Model}
\begin{figure}[ht]
\begin{center}
\includegraphics[width=\columnwidth, height=0.3\textheight]{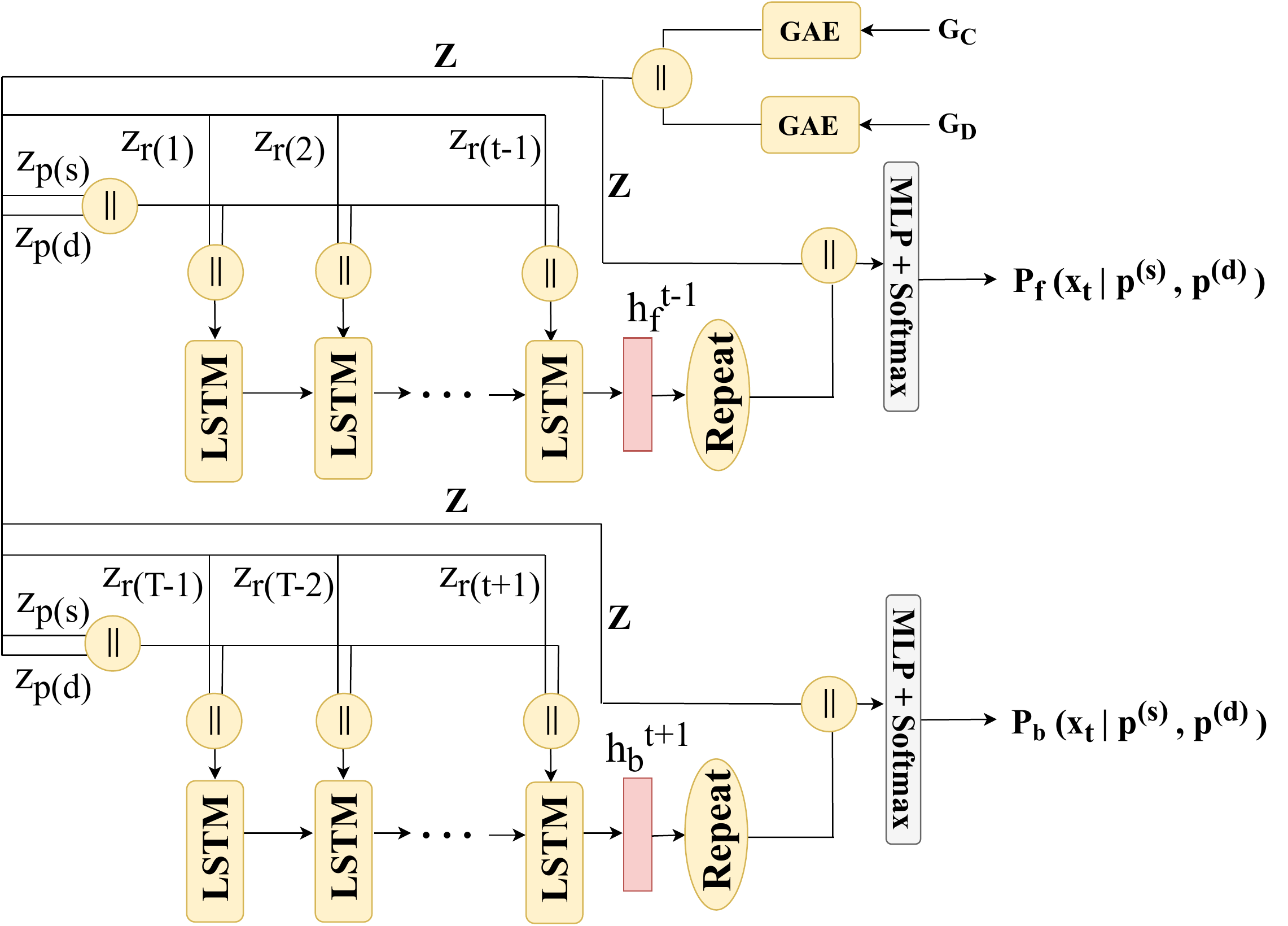}
\caption{The Itinerary Net (ITRNet)}
\label{ITRNet}
\end{center}
\end{figure}

The ITRNet consists of two LSTM's, namely a \textit{forward} and a \textit{backward} LSTM. The forward LSTM takes a partial route sequence from the starting POI and estimates the probability of a POI being the next POI in the route sequence. The backward LSTM takes a route sequence in reverse order, starting from the destination POI, to estimate the next POI in the reverse route sequence. Both LSTM's also take into consideration the actual source and destination POIs of the route being generated.

Let $R =(r_1,r_2,...,r_{t-1},x_t,r_{t+1}...,r_T)$ be a route, $p^{(s)} = r_1$ and $p^{(d)} = r_T$ are the source and destination POIs respectively. Formally,  given the forward partial sequence $r_{1:t-1}$, the source $p^{(s)}$ and destination $p^{(d)}$, the probability of $i^{th}$ POI $p_i$ $\in Q$ replacing $x_t$,, can be computed as:
\begin{equation}\label{equation_pf}
    {P_f}(x_{i,t}|p^{(s)},p^{(d)}) = p(x_t = p_i|r_{1:t-1},p^{(s)},p^{(d)})
\end{equation}
Similarly, given the backward partial sequence $r_{t+1:T}$, the source $p^{(s)}$ and destination $p^{(d)}$, the probability of $i^{th}$ POI $p_i \in Q$ replacing $x_t$ can be computed as:
\begin{equation}\label{equation_pb}
    {P_b}(x_{i,t}|p^{(s)},p^{(d)}) = p(x_t = p_i|r_{t+1:T}, p^{(s)},p^{(d)})
\end{equation}

Similarly, we compute the probabilities of all $N$ POIs in $Q$, which is denoted as $N$-dimensional vectors $\mathbf{P_f}(x_t|p^{(s)},p^{(d)})$ and $\mathbf{P_b}(x_t|p^{(s)},p^{(d)})$ respectively, where each element $i$ in $\mathbf{P_f}(x_t|p^{(s)},p^{(d)})$ and $\mathbf{P_b}(x_t|p^{(s)},p^{(d)})$  represents the conditional probabilities defined in Equation \ref{equation_pf} and \ref{equation_pb} respectively. The forward LSTM computes $\mathbf{P_f}(x_t|p^{(s)},p^{(d)})$, while the backward LSTM computes $\mathbf{P_b}(x_t|p^{(s)},p^{(d)})$.

To develop the ITRNet, we first compute POI embeddings, which would be later used to train the subsequent ITRNet forward and backward LSTM models.

%
\subsubsection{POI Embedding}
To capture the spatial and semantic information of POIs in the ITRNet, we use two graph autoencoders \cite{kipf2016variational} to get the embeddings of POIs. These embedding enable the LSTM models to accurately compute the probability of POIs even if the historical visits to the POIs are sparse.

We first define two graphs $G_C = (Q,E_C)$ and $G_D = (Q,E_D)$ for POI categories and spatial distance, respectively and the nodes of both graphs are the POIs. The weight of an edge between two nodes in $G_C$ and $G_D$ is related to the categorical similarity and the distance of the two corresponding POIs, respectively. 

The adjacency matrix $A_C$ for graph $G_C$ is defined as follows.
\begin{displaymath}
A_{C_{ij}} = \left\{
\begin{array}{cc}
	1 & \textrm{if }  p_i.cat = p_j.cat\\
	0 & \textrm{otherwise } \\
\end{array} 
\right.
\end{displaymath}

This adjacency matrix captures the categorical similarity between POIs, where POIs with the same category have a connection. On the other hand, the adjacency matrix $A_D$ of $G_D$ is defined as follows.
$$ A_{D_{ij}} = \frac{e^{d_{max} - dist(p_i,p_j)} - e^{d_{max} - d_{min}}}{1 - e^{d_{max} - d_{min}}} $$
Here $dist(p_i,p_j)$ is the distance between POI $p_i.loc$ and $p_j.loc$, and $d_{max}$ and $d_{min}$ are the maximum and minimum distances between any two POIs, respectively.  

We use the Euclidean distance between two POIs calculated using their latitude and longitude, which is a reasonable approximation of the road network distance \cite{hua2018euclidean}. Without loss of generality, Euclidean distances can be replaced with road network distances if underlying road network is available. We use exponential terms to amplify the difference between the normalized weight values. The $A_D$ matrix gives a larger weight to edges between POIs that are nearer to each other. For the lowest distance $d_{min}$ between any two nodes $p_i$ and $p_j$, the edge weight $A_{D_{ij}}$ between them will be 1. On the other hand the two nodes with the maximum distance $d_{max}$ between them will have an edge weight of 0.

We obtain two embeddings $\mathbf{Z_C}\in\mathbb{R}^{|Q| \times e_c}$ and $\mathbf{Z_D}~\in~\mathbb{R}^{|Q| \times e_d}$ from the two graph autoencoders respectively, where $e_c$ and $e_d$ are the embedding dimensions. We concatenate these two embeddings to obtain the final POI embeddings, $\mathbf{Z} = \mathbf{Z_C} || \mathbf{Z_D}$, where $\textbf{Z}~\in~\mathbb{R}^{|Q| \times (e_d+e_c)}$,`$||$' is the concatenation operation and $i^{th}$ row of $\mathbf{Z}$ corresponds to the embedding of POI $p_i~\in~Q$.

The reconstructed adjacency matrix for $A$ is calculated as:
$$\hat{A} = RELU\left( \mathbf{ZZ^{T}}\right)$$ for both autoencoders, where $\mathbf{Z} = GCN(\mathbf{X},\mathbf{A})$. Here, $\mathbf{A}$ is the POI ground truth adjacency matrix, $\mathbf{X}$ is the featureless identity matrix input, and GCN is a graph convolution operation~\cite{kipf2016semi}. The learned node embeddings $\mathbf{Z_D}$ thus captures the categorical and distance information between POIs, which assists the downstream LSTM models to predict the next POI in a route sequence more accurately. We use MSE and cross-entropy loss functions for the autoencoders corresponding to $G_D$ and $G_C$, respectively.

\subsubsection{Forward and Backward LSTM Models}

This is the main component of ITRNet, which uses two LSTMs to generate two conditional probabilities based on a known partial route sequence, a given source POI and a given destination POI. It also uses the POI embeddings obtained from the previous step.


To estimate the \textit{forward} conditional probability $P_f(x_{i,t}|p^{(s)},p^{(d)})$, we first obtain the encoding of the given partial route sequence. The encoding of a sub-route upto any position $j$, that is the partial route $(r_1,r_2,...,r_j)$ where $1 \leq j \leq t-1$ is obtained using a forward LSTM model as follows.

$$\mathbf{h}_{f}^{j} = LSTM_F( \mathbf{z}_{r_j} || \mathbf{z}_{p^{(s)}} || \mathbf{z}_{p^{(d)}}, \mathbf{h}_{f}^{j-1})$$
where, $\mathbf{z}_{r_j}$, $\mathbf{z}_{p^{(s)}}$, $\mathbf{z}_{p^{(d)}}$ is the embedding of POI $r_j$ at step $j$, source POI $p^{(s)}$ and destination POI $p^{(d)}$ respectively, $\mathbf{h}_{f}^{j-1}$ is the hidden state LSTM vector at $j-1$, and $\mathbf{h}_{f}^{j}$ is the hidden state vector at step $j$.

After obtaining the encoding of the observed sub-route, we calculate the probability of a POI $p_i$ for position $t$ as follows.
$$ P_f(x_{i,t}|p^{(s)},p^{(d)})  = \frac{\alpha_{i,t}}{ \sum_{i=1}^{|N|} \alpha_{i,t}}$$
where
$$ \alpha_{i,t} = MLP(\mathbf{z}_{p_i} || \mathbf{h}_{f}^{t-1}) $$
Here $\mathbf{z}_{p_i}$ is the POI embedding of POI $p_i$ , $\mathbf{h}_{f}^{t-1}$ is the sub-route encoding upto $t-1$ and $MLP$ is a two-layer perceptron network, which outputs a score $\alpha_{i,t}~\in~\mathbb{R}$. Passing these scores through a softmax layer gives us the final forward conditional probability estimation vector $\mathbf{P_f}(x_t|p^{(s)},p^{(d)})$. The $MLP$ thus computes the probability of a POI $p_i$ being the next POI $r_t$ in the route, given the encoding of the partial route upto step $t-1$ and the embedding of the POI $\mathbf {z}_{p_i}$.

Similarly, we develop the \textit{backward} LSTM model to estimate the backward conditional probability $P_b(x_{i,t}|p^{(s)},p^{(d)})$. It takes the backward sub-route $(r_{t+1}, r_{t+2},...,r_T)$ in \textit{reverse} order, along with the source POI $p^{(s)}$ and destination POI $p^{(d)}$. Essentially, we generate the encoding $\mathbf{h}_{b}^{j}$ at step $j$, where $t+1 \leq j \leq T$, as follows. 
$$\mathbf{h}_{b}^{j} = LSTM_B(\mathbf{z}_{r_j} || \mathbf{z}_{p^{(s)}} || \mathbf{z}_{p^{(d)}}, \mathbf{h}_{b}^{j+1})$$
Using the above encoding and the earlier POI embeddings we estimate $\mathbf{P_b}(x_t|p^{(s)},p^{(d)})$. The procedure is similar to the procedure to obtain $\mathbf{P_f}(x_t|p^{(s)},p^{(d)})$, so we do not repeat it here. Thus the forward LSTM predicts the next POI given a forward partial route sequence, whereas the backward LSTM predicts the next POI in the reverse partial route sequence, i.e. the immediate previous POI given a sequence of POIs visited after the predicted POI.

While training, we adopt the binary cross-entropy loss to train both the LSTM models.

\subsection{Generating Itineraries Using ITRNet}

In this phase, we generate $k$ alternative itinerary recommendations using the ITRNet backward and forward LSTM models. Given a query $q(s,d,k)$, we first compute a \textit{relevancy} score of all POIs for a given source POI $s$ and destination POI $d$ using the ITRNet. Then, at each iteration we extract a \textit{prominent POI} based on the computed relevancy scores and generate $k$ alternative itineraries each going through a different prominent POI. We first describe how we compute the POI relevancy, after that we describe how an itinerary is generated. Finally we describe how $k$ alternative itineraries are obtained in an iterative manner.

\subsubsection{Computing POI Relevancy}
By using the ITRNet, we define a relevancy function that outputs a relevancy score for every POI for given query source POI $s$ and destination POI $d$. 

Consider the route  $(r_1 = s,r_2 = x_2,r_3 = d)$, where $x_2$ is the variable POI. We define the $relevancy$ function as follows:
$$relevancy(p_i) = \frac{1}{2}({P_f}(x_{i,2}|s,d) + {P_b}(x_{i,2}|s,d))$$
This $relevancy$ function provides higher scores to POIs which are more relevant in the context of the query source and destination POIs.

\subsubsection{Generating an Itinerary}\label{subsub:generate_itinerary}
Based on the relevancy function, we obtain a \textit{prominent POI}, $p^{(p)}$, the POI with the maximum relevancy score. After obtaining a prominent POI, we generate an itinerary containing that prominent POI. \methodname~ uses the forward and backward LSTM models of ITRNet to generate the partial itinerary from the prominent POI to the source POI and the partial itinerary from the prominent POI to the destination POI. We call the partial itinerary from the source POI to the prominent POI the \textit{first half itinerary} and the partial itinerary from the prominent POI to the destination POI the \textit{second half itinerary}. 

We generate the half itineraries starting from the prominent POI $p^{(p)}$, as then the corresponding LSTM that would be used will output probabilities with the knowledge that the prominent POI is present in the itinerary being generated.

There are two ways to develop the full itinerary through generating the first and second half itineraries starting from the prominent POI:
\begin{itemize}
    \item Generate the first half itinerary in reverse order using the backward LSTM model of ITRNet. Then given the first half itinerary as partial sequence, generate the second half itinerary using the forward LSTM model.
    \item Generate the second half itinerary using the forward LSTM model. Then given the second half itinerary as a partial sequence, generate the first half itinerary in reverse order using the backward LSTM model.
\end{itemize}

Note that in all cases the source and destination POI input to the LSTM models are the query source and destination POIs $s$ and $d$, respectively. We assume a maximum allowable length of a half itinerary $L_{max}$. To obtain the first half itinerary (following the first way), we use the backward LSTM of ITRNet. We place the prominent POI at position $L_{max}$ and generate POI probabilities $\mathbf {P_b}(x_t|s,d)$ for positions $L_{max} - 1$ to $1$. We determine the position of the source POI in the reverse first half itinerary as:
$$ t_s = \argmax_{1 \leq t \leq L_{max} - 1} {P_b}(x_{s,t}|s,d) $$
We place the source POI in position $t_s$. For all other positions from $L_{max} - 1$ down to $t_s$ we choose the POI $r_{t}$ in the sequence through:
$$r_{t} = \argmax_{p_i} {P_b}(x_{i,t}|s,d)$$
Note that during this choice we avoid selection of a POI which is already in the partial sequence generated to avoid loops in the recommended itinerary. We also avoid selection of the given source and destination POIs too. Finally we adopt the partial itinerary sequence from position $t_s$ to $L_{max}$ as our first half itinerary.

After generating the first half itinerary, we generate the second half itinerary from position $L_{max} + 1$ to $2L_{max}$ given the first half itinerary, source POI $s$ and destination POI $d$. We determine the position of the destination POI in the second half itinerary as:
$$ t_d = \argmax_{L_{max}+1 \leq t \leq 2L_{max}} {P_b}(x_{d,t}|s,d) $$
We place the destination POI at $t_d$. For all other positions from $L_{max} + 1$ to $t_d$ we choose the POI $r_{t}$ in the sequence using the forward LSTM model through:
$$ r_{t} = \argmax_{p_i} {P_f}(x_{i,t}|s,d)$$
Finally we put $d$ at position $t_d$. The partial itinerary sequence from $L_{max} + 1$ to $t_d$ make up our second half itinerary. Thus the first and second half itineraries make up our desired itinerary from POI $s$ to $d$ through the given prominent POI $p^{(p)}$.

Similarly we can generate the itinerary using the second way as mentioned above. We place the prominent POI $p^{(p)}$ at position $L_{max}$. Among the two generated itineraries, we choose the one with the lowest \textit{perplexity} or negative log likelihood, calculated using the forward LSTM model:
\begin{equation}\label{equation_perplexity}
    perplexity(I) = - \sum_{i=1}^{|I|} \log {P_f}(x_{r_i,i}|s,d)
\end{equation}
Where $I$ is the itinerary $(r_1,r_2,\cdots, r_{|I|})$ of length ${|I|}$ with $r_1 = s$ and $r_{|I|} = d$. 


\subsubsection{Generating $k$ Alternative Itineraries}
To obtain $k$ alternative itineraries as specified in a given query $q(s,d,k)$, we generate itineraries iteratively through determining a prominent POI at each iteration. We keep track of the total number of occurrences for all POIs in the itineraries generated until the current iteration. Let $O[p_i]$ be the number of occurrences of POI $p_i$ in all the itineraries generated up to the current iteration.
At each iteration $j$, we obtain the set of POIs with minimum occurrence $O_{min} = \{p_i | O[p_i] = \min_{p_i} O[P_i]\}$. Then we obtain the prominent POI $p^{(p)}$ at any iteration as:
\begin{equation}\label{ppoi_eq}
     p^{(p)} = \argmax_{p_i \in O_{min}}~~~relevancy(p_i)
\end{equation}

We then take $p^{(p)}$ and generate an itinerary as described in Section \ref{subsub:generate_itinerary} using $p^{(p)}$ as the prominent POI. After obtaining an itinerary we update the values of $O[p_i]$ for each $p_i$ in the itinerary obtained in this iteration.

We run the same process $k$ times and obtain our desired $k$ itineraries. An overview of the whole \methodname system is given in Algorithm \ref{Algo}.
\begin{figure}[!t]
 \removelatexerror
\begin{algorithm}[H]
\caption{\methodname}
\label{Algo}
\SetAlgoLined
\DontPrintSemicolon
\SetKwInOut{Input}{Input }
\SetKwInOut{Output}{Output }
set $O[p_i] = 0$ for every $p_i\in Q$\;
$I \gets \emptyset $\;
\For{$i = 1,\dots,k$}{
$p^{(p)} \gets$ obtain prominent POI using Equation~\ref{ppoi_eq}\;
$I_i \gets $ generate itinerary passing through $p^{(p)}$\;
$I \gets I \cup I_i$\;

set $O[p_i] = O[p_i] + 1$ for every $p_i\in I_i$ 

}
Return $I$
\end{algorithm}
\end{figure}
\subsection{Generating Itineraries Through Sampling}\label{sub:sampling_algorithm}


In a real-world scenario, a user may want to impose some constraints on the generated alternative itineraries, such as setting a fixed budget or time limit, or specifying must-see POIs, etc. It is not possible to support such constraints in our proposed LSTM based trip generation technique described in Section \ref{subsub:generate_itinerary}. To overcome such limitations, in this section, we propose an alternate sampling algorithm to generate an itinerary starting from $p^{(s)}$, passing through the prominent POI $p^{(p)}$, and ending at $p^{(d)}$. Our sampling based approach is as follows.

We iteratively generate candidate itineraries. At iteration $0$ of the sampling method, we start with an initial itinerary $R^{(0)} = (r_1^{0}=s, r_1^{0}=p^{(p)}, r_2^{0}=d)$. In the iteration process, at iteration $j$, suppose we have itinerary $R^{(j)} = (r_1^{(j)}, r_2^{(j)} , ..., r_{{|I|}_{j}}^{(j)})$ of length ${|I|}_j$. We generate a sample at iteration $j+1$, i.e., 
$R^{(j+1)} = (r_1^{(j+1)}, r_2^{(j+1)} , ..., r_{{|I|}_{j+1}}^{(j+1)})$ of length ${|I|}_{j+1}$ by modifying sample $R^{(j)}$. 

For modification, we randomly select a POI at position $t$ of the itinerary $R^{(j)}$, where $2 \leq t \leq {|I|}_{j} - 1$. Then we perform one of the following four operations at $t$, namely at POI $r_t^{(j)}$ of itinerary $R^{(j)}$:

\underline{\textit{Insertion:}} We first assign $r_t^{(j+1)} = r_t^{(j)}$. We then insert a POI between $t$ and $t+1$. We now define the following conditional probability distribution using the ITRNet:
$$ {P_c}(x_{i,t}|p^{(s)},p^{(d)}) = {P}(x_{i,t}|r_{1:t-1},r_{t+1:T}, p^{(s)}, p^{(d)})$$
The above equation gives us the probability of a POI at a given position given both the partial sequence from the source POI to the POI before $p_i$ at position $t-1$ and the partial sequence starting after $p_i$ from position $t+1$ to the destination POI at position $T$. We can compute this for all POIs in $Q$, which can be denoted as an $N$ dimensional vector $\mathbf {P_c}(x_t|p^{(s)},p^{(d)})$. We compute $\mathbf {P_c}(x_t|p^{(s)},p^{(d)})$ as follows:
\begin{multline}\label{eqn:prob_combined}
    \mathbf{P_c}(x_t|p^{(s)},p^{(d)}) = \beta * \mathbf{P_f}(x_t|p^{(s)},p^{(d)}) \\ + (1 - \beta) * \mathbf{P_b}(x_t|p^{(s)},p^{(d)})
\end{multline}
where, $\beta = \frac{t}{T-1}$. Intuitively, we give more weights to the model that has seen a longer sub-route and thus have a greater contextual information.

We obtain $\mathbf {P_c}(x_{t+1}|s,d)$ at position $t+1$, where the newly inserted POI is located in the itinerary. We sample a POI from $\mathbf {P_c}(x_{t+1}|s,d)$ and assign the obtained sample as $r_{t+1}^{(j+1)}$. The rest of the itinerary remains unchanged.

\underline{\textit{Deletion}}: We delete $r_t^{(j)}$ at position $t$ of the itinerary and keep the rest of the itinerary unchanged.

\underline{\textit{Replacement}}: We obtain the conditional probability distribution $\mathbf{P_c}(x_t|p^{(s)},p^{(d)})$ as in Equation \ref{eqn:prob_combined} using the ITRNet. We then take a sample from this distribution to obtain a POI $p_r$ and get $r_t^{(j+1)} = p_r$ replacing POI $r_t^{(j)}$.

\underline{\textit{Swap and Replace}}: We randomly select a position $t_{rand}$, between $2$ to ${|I|}_{j} - 1$, and swap the position of POI $r_t^{(j)}$ at position $t$ and POI $r_{t_{rand}}^{(j)}$ at position $t_{rand}$. If $r_{t_{rand}}^{(j)}$ is not a prominent POI, we also perform the \textit{Replacement} operation (as described in the previous paragraph) at position $t$ after the swap.

To perform an operation at any iteration $j$, we choose any one of the operations with equal probability assigned to all allowed operations. If the selected POI $r_t^{(j)}$ is the prominent POI, we do not perform the deletion or replacement operation on that POI. Also, to avoid loops, we omit inserting or putting through replacement a POI that is already present in $R^{(j)}$ except for POI $r_t^{(j)}$.

At each iteration we check the following two conditions:
\begin{itemize}
    \item All the pre-defined user constraints are satisfied (if any)
    \item The \textit{perplexity} of $R^{(j+1)}$ as defined in Equation \ref{equation_perplexity} is lower than itinerary $R^{(j)}$ at iteration $j$, or no new itinerary has been accepted for previous two iterations.
\end{itemize}
If both conditions are satisfied, we adopt itinerary $R^{(j+1)}$ for generating itinerary at iteration $j+2$. Otherwise we retain itinerary $R^{(j)}$ and use this to generate itinerary at iteration $j+2$, meaning we reject the modification operation performed at iteration $j+1$. The sampling runs for $J$ iterations. The itinerary generated and accepted with the minimum perplexity is returned as the desired itinerary. Note that if any user defined constrains is given, we have to first build an initial itinerary satisfying all the given constraints. Any such itinerary that satisfies all the conditions given will suffice as the initial itinerary.
\subsubsection{Satisfying User Constraints}
Our sampling algorithm makes it possible to generate alternative  itineraries that can satisfy a variety of user constraints. Examples include:
\begin{itemize}
    \item  \textit{A given fixed budget}: If the cost from visiting one POI to another is given, users may want itineraries that they can visit within a fixed budget. We may omit a candidate itinerary generated at an iteration if the itinerary exceed the budget.
    \item \textit{Must see POIs}: Users may want itineraries that must include one or more specific given POIs. In such cases, we keep those POIs in the initial sequence and treat them similar to the prominent POI, i.e., we don't delete or replace those POIs.
    \item \textit{Time constraints for POIs and Itineraries}: Many times POIs have opening and closing hours. Given a start time along with the source and destination POIs, the average staying time in a POI and average travel times between POIs, we can check whether all the constraints are met while generating itineraries in different iterations. We can also consider only those POIs during sampling in insertion or replacement that would satisfy the time constraints. Also users may want itineraries that they can travel within a fixed time limit. This can be also satisfied, where we omit itineraries generated in an itinerary when the time budget is not satisfied.
\end{itemize}
Note that the aforementioned constraints cannot be trivially satisfied in a traditional deep learning algorithm. Thus the itinerary generation technique of \textit{\methodname-Samp} is effective in many practical scenarios for generating itineraries in a constraint setting.
\section{Experiments}
In this section, we present the experimental evaluations for \emph{\methodname} to recommend $k$ alternative itineraries for a given source and destination POI. In particular, depending upon the itinerary generation strategy we have two versions of \emph{\methodname}: (i) \emph{\methodname-LSTM} that uses LSTMs for generating an itinerary (Section~\ref{subsub:generate_itinerary}), and (ii) \emph{\methodname-Samp} that adopts a sampling based flexible approach for generating an itinerary (Section~\ref{sub:sampling_algorithm}). 


\subsection{Baselines}
We are the first to \textit{learn} alternative itineraries from historical routes. As there are no prior works in the literature that directly solves our problem, we adapt two state-of-the-art trip recommendation techniques that learn from historical routes, and modify them to recommend $k$ alternative itineraries. Our two baselines are as follows.
\begin{itemize}
\item \textit{Markov+DBS:} We extend the \textit{Rank+Markov} of \cite{chen2016learning} to generate $k$ alternative itineraries. We first compute score ranks for POIs based on their features for a given query. Then a POI-to-POI transition matrix is computed from feature-to-feature transition probabilities. From the computed POI scores and transition probabilities, we use Viterbi algorithm to generate a route of a specific length for a given  source and destination. We incorporate the diversified beam search measure given by \cite{li2016simple} to maintain $k$ paths at each step of the algorithm.

\item \textit{NASR+DBS:} We adopt NASR~\cite{wang2019empowering} which uses self-attention-based LSTM to estimate the conditional probability (similar to our forward LSTM model). Again we run the diversified beam search \cite{li2016simple} on top of this model and return the top $k$ itineraries ending at the destination POI with the highest probability scores.
\end{itemize}

We evaluate the effect of a number of parameters. Specifically, the no. of alternative itineraries recommended, $k$, is varied from $1$ to $5$ with a default value of $3$. Also to ensure fair comparison, we fix the length $L$ of each itinerary (i.e., number of POIs in it including $s$ and $d$) recommended by different algorithms. $L$ is varied from $3$ to $9$ with the default value being $5$.
We use a 5-fold cross validation: one fold is kept for testing and the other four folds are used to train a model. The average performance metrics among all five folds are reported.

\subsection{Datasets}
We use eight popular real-world datasets drawn from two different domains. As a first group of datasets, we take geo-tagged Flickr traces of three touristic cities: Edinburgh, Toronto and Melbourne \cite{lim2015personalized, chen2016learning}. In the second group of datasets, we consider trips of five different theme parks:  California Adventure, Hollywood, Disneyland, Disney Epcot and Magic Kingdom~\cite{lim2017personalized}. 

Along with the trips involving different POIs, the datasets also contain location $\langle lat, long\rangle$ and the category of each POI. The trajectories given in these datasets are generated from user check-ins, with the visiting time between two consecutive POIs in a trajectory is no more than 8 hours.
We filter out multiple occurrences of POIs (if any) from these trajectories to avoid loops. We also only consider trajectories having at least three POIs.
Table~\ref{Dataset_table} shows the details of each dataset including the number of POIs, number of routes having at least three POIs and the number of ground truth set of routes generated with unique ($s$, $d$) pairs.
\begin{table}[ht]
\begin{center}
\resizebox{\columnwidth}{!}{
\begin{tabular}{|l|c|c|c|}
    \hline
    \textbf{Place} & \textbf{\# POIs} & \textbf{\# routes} & \textbf{unique ($s,d$) pairs} \\
    \hline \hline
 Edinburgh & 28 & 634 & 267\\ \hline
     Toronto & 29 & 335 & 163 \\    \hline
     Melbourne & 88 & 442 & 373 \\
    \hline \hline
     California Adventure & 25 & 1475 & 404\\ \hline
     Hollywood & 13 & 901 & 134 \\ \hline
    
     Disneyland & 31 & 2792 & 618 \\ \hline
     Disney Epcot & 17 & 1248 & 207 \\ \hline
     Magic Kingdom & 27 & 2218 & 508 \\ \hline
\end{tabular}
}
\end{center}
\caption{\label{Dataset_table}Dataset Statistics}
\end{table}
\begin{table*}[ht]
\begin{subtable}[t]{2\columnwidth}
\resizebox{\columnwidth}{!}{
\begin{tabular}[t]{|c|c|c|c|c|c|c|c|c|c|c|c|c|}
\hline
   & \multicolumn{3}{|c|}{\textbf{Edinburgh}} & \multicolumn{3}{|c|}{\textbf{Toronto}} & \multicolumn{3}{|c|}{\textbf{Melbourne}} & \multicolumn{3}{|c|}{\textbf{CaliAdv.}} \\
\hline
   \textbf{Method} & F1 & Pairs-F1 & Div. & F1 & Pairs-F1 & Div. & F1 & Pairs-F1 & Div. &F1 & Pairs-F1 & Div.\\
\hline
   Markov+DBS & \textbf{0.592} & \textbf{0.287} & 0.534 & \textbf{0.597} & 0.284 & 0.461 & 0.500 & 0.186 & 0.564 & 0.527 & 0.216 & 0.478 \\
\hline
   NASR+DBS & 0.576 & 0.268 & 0.643 & 0.583 & 0.267 & 0.666 & 0.497 & 0.181 & 0.722 & \textbf{0.541} & \textbf{0.229} & 0.576 \\
\hline
   \methodname-LSTM & 0.580 & 0.272 & \textbf{0.766} & 0.589 & 0.280 & \textbf{0.755} & \textbf{0.521} & \textbf{0.207} & 0.791 & 0.531 & 0.221 & \textbf{0.771}\\
\hline
   \methodname-Samp & 0.577 & 0.270 & 0.724 & 0.594 & \textbf{0.286} & 0.665 & 0.512 & 0.198 & \textbf{0.814} & 0.526 & 0.214 & 0.766\\
\hline
\end{tabular}
       }
\caption{}
\end{subtable}

\bigskip 

\begin{subtable}[t]{2\columnwidth}
\resizebox{\columnwidth}{!}{
\begin{tabular}[t]{|c|c|c|c|c|c|c|c|c|c|c|c|c|}
\hline
   & \multicolumn{3}{|c|}{\textbf{DisHolly}} & \multicolumn{3}{|c|}{\textbf{Disland}} & \multicolumn{3}{|c|}{\textbf{Epcot}} & \multicolumn{3}{|c|}{\textbf{MagicK}} \\
\hline
   \textbf{Method} & F1 & Pairs-F1 & Div. & F1 & Pairs-F1 & Div. & F1 & Pairs-F1 & Div. &F1 & Pairs-F1 & Div.\\
\hline
   Markov+DBS & 0.607 & 0.297 & 0.343 & 0.513 & 0.203 & 0.383 & 0.585 & 0.279 & 0.425 & 0.512 & 0.205 & 0.394 \\
\hline
   NASR+DBS & \textbf{0.631} & \textbf{0.329} & 0.478 & \textbf{0.533} & \textbf{0.222} & 0.497 & \textbf{0.604} & \textbf{0.298} & 0.422 & \textbf{0.525} & \textbf{0.215} & 0.525 \\
\hline
   \methodname-LSTM & 0.623 & 0.319 & 0.572 & 0.515 & 0.206 & 0.733 & 0.585 & 0.280 & \textbf{0.741} & 0.515 & 0.205 & \textbf{0.761}\\
\hline
   \methodname-Samp & 0.615 & 0.315 & \textbf{0.668} & 0.506 & 0.197 & \textbf{0.796} & 0.582 & 0.276 & 0.704 & 0.509 & 0.203 & 0.761\\
\hline
\end{tabular}
       }
\caption{}
\end{subtable}

\caption{Comparison w.r.t. F1, Pairs-F1 and Diversity Scores}
\label{tab:table_f1_pf1_div_all}
\end{table*}
\subsection{Evaluation Metrics}\label{sub:eval_metrics}
Given a query $q(s,d,k)$, we use $S_{REC}$ to denote the set of $k$ recommended itineraries by an algorithm and $S_{gr}$ to denote the ground truth routes which consists of all the historical routes that start at $s$ and end at $d$. Next, we describe the metrics that we use to measure the quality of itineraries $S_{REC}$ returned by an algorithm. In particular, we measure the quality of our alternate itineraries by using traditional popularity and diversity metrics independently as well as by another metric that considers both popularity and diversity at the same time. 


\subsubsection{Popularity}\label{subsub:popularity_metric}


We we use the widely used F1 score and pairs-F1 score \cite{lim2015personalized,chen2016learning} to measure the popularity $pop(S_{REC})$ of a set of $k$ recommended itineraries $S_{REC}$.

Suppose, route $R = (r_1,r_2,r_3,...,r_{|R|})$ is a ground truth route where $r_1=s$ and $r_{|R|} = d$ and itinerary $I= (i_1,i_2,...,i_|I|)$ is a recommended itinerary. Also, let $Q_R$ and $Q_I$ be the sets of POIs in the ground truth route and recommended itinerary respectively. The precision of the recommended itinerary $|I|$ is calculated as:$ P = \frac{|Q_R \cap Q_I|}{|Q_I|} $ and the recall is calculated as $R = \frac{|Q_R \cap Q_I|}{|Q_R|}$. The F1 score is the harmonic mean of precision and recall, i.e. $F1 = \frac{2PR}{P + R}$.

In contrast to F1, pairs-F1 score considers orders of POIs in the routes. Specifically, precision is  the no. of ordered POI pairs present in both $R$ and $I$ divided by the total no. of ordered POI pairs in $I$. The recall is the total no. of ordered POI pairs present in both $R$ and $I$ divided by the total number of ordered POI pairs in $R$.
 The pairs-F1 score is the harmonic mean of this precision and recall.

We compute F1 score (resp. pairs-F1 score) for each pair in $S_{REC}\times S_{gr}$ and report the average value as the popularity of the recommended itinerary set $S_{REC}$. The popularity scores indicate the average quality of the individual recommended itineraries with respect to the historical trips adopted by past users. Note that the F1 or pairs-F1 scores are always between 0 and 1.

\subsubsection{Diversity}\label{subsub:diversity_metric}
We adapt the diversity metric used in~\cite{benouaret2016package}, originally defined to measure the diversity among POIs in a set.
To measure the diversity of a set of itineraries $S_{REC}$ containing $k$ itineraries, we first measure a \textit{similarity} value $sim(I_i,I_j)$ for a pair of itineraries $I_i$ and $I_j$. We then define the diversity value of a recommended set of itineraries $S_{REC} = \{I_1, I_2, \cdots, I_k\}$ of size $k$ as 


\begin{equation}\label{eqn:diversity_formula}
    div(S_{REC}) = \frac{1}{k(k-1)}\sum_{\substack{(I_i,I_j)\in S_{REC}\times S_{REC} \\ i \neq j}}(1 - sim(I_i,I_j))
\end{equation}

In Equation \ref{eqn:diversity_formula}, we calculate the average diversity between all $k(k-1)$ pairs in the recommended itinerary set $S_{REC}$. For each pair, the diversity value is the dissimilarity value between the two itineraries, i.e., $1 - sim(I_i,I_j)$. As long as the similarity measure $sim(I_i,I_j)$ is between 0 to 1, the dissimilarity measure and thus $div(S_{REC})$ will also remain between 0 to 1, with higher value indicating higher diversity between the recommended itineraries. In our experiments, we adopt the F1 score between a pair of itineraries as the similarity measurement, ignoring the source and destination POIs.

Also observe that, the notion of popularity and diversity is somewhat conflicting. For example, suppose a model is to recommend 2 alternative itineraries. If the model recommends the most popular itinerary 2 times, it will achieve the maximum average popularity score, but the diversity score of the recommended itinerary set would be 0. If it attempts to diverse from this most popular itinerary, it would achieve the diversity score. But unless it can align with an alternatively popular itinerary, the popularity measure of this second recommended itinerary will be low and the average popularity score will drop. Thus to achieve higher popularity and diversity scores at the same time, a model must recommend quality alternative itineraries with respect to the historical trips.

\subsubsection{Combination of Popularity and Diversity}
Since our goal is to recommend quality alternative itineraries that are both popular and diverse, we need a measure that combines the popularity and diversity of $S_{REC}$. We employ the widely used weighted sum to define the combined metric as $$comb(S_{REC}) = \alpha \times pop(S_{REC}) + (1 - \alpha) \times div(S_{REC})$$ where $pop(S_{REC})$ is the average popularity score of the recommended itineraries which is computed using the F1 score as discussed earlier in Section \ref{subsub:popularity_metric}.
The parameter $\alpha \in [0,1]$ specifies the relative importance of popularity and diversity.For example, $\alpha < 0.1$ gives very little importance to the popularity of the itineraries, and $\alpha > 0.9$ provides very little importance to the diversity of the itineraries. Thus we consider $0.1 \leq \alpha \leq 0.9$ during our evaluation. To give equal importance to both popularity and diversity, we set $\alpha = 0.5$ as a default value. 



It is important to note that our proposed approaches, both \textit{\methodname-LSTM} and \textit{\methodname-Samp} are agnostic to the above metrics, and our motivation is to \textit{learn} popular alternative routes without any such explicit modeling of popularity and/or diversity. Yet, we show that that our proposed learning-based approaches outperform baselines significantly w.r.t. these traditional metrics. 

\begin{table*}[ht]
\begin{subtable}[t]{2\columnwidth}
\resizebox{\columnwidth}{!}{
\begin{tabular}[t]{|c|c|c|c|c|c|c|c|c|c|c|c|c|}
\hline
   & \multicolumn{4}{|c|}{\textbf{F1}} &  \multicolumn{4}{|c|}{\textbf{Diversity}} & \multicolumn{4}{|c|}{\textbf{Comb ($\boldsymbol{\alpha}$ = 0.5)}} \\
\hline
   \textbf{L} & 3 & 5 & 7 & 9 & 3 & 5 & 7 & 9 & 3 & 5 & 7 & 9\\
\hline
   Markov+DBS & 0.631 & \textbf{0.592} & \textbf{0.542} & \textbf{0.503} & 1.000 & 0.534 & 0.240 & 0.130 & 0.816 & 0.563 & 0.391 & 0.316 \\
\hline
   NASR+DBS & 0.632 & 0.576 & 0.525 & 0.480 & 1.000 & 0.643 & 0.408 & 0.232 & 0.816 & 0.610 & 0.467 & 0.356 \\
\hline
   \methodname-LSTM & \textbf{0.644} & 0.580 & 0.521 & 0.480 & 1.000 & \textbf{0.766} & \textbf{0.632} & \textbf{0.548} & \textbf{0.822} & \textbf{0.673} & \textbf{0.576} & \textbf{0.514} \\
\hline
   \methodname-Samp & 0.644 & 0.577 & 0.521 & 0.476 & 1.000 & 0.724 & 0.592 & 0.482 & 0.822 & 0.651 & 0.556 & 0.479\\
   

\hline
\end{tabular}
       }
\label{table_edin_L}
\caption{Edinburgh Dataset}
\end{subtable}

\bigskip 

\begin{subtable}[t]{2\columnwidth}
\resizebox{\columnwidth}{!}{
\begin{tabular}[t]{|c|c|c|c|c|c|c|c|c|c|c|c|c|}
\hline
   & \multicolumn{4}{|c|}{\textbf{F1}} &  \multicolumn{4}{|c|}{\textbf{Diversity}} & \multicolumn{4}{|c|}{\textbf{Comb ($\boldsymbol{\alpha}$ = 0.5)}} \\
\hline
   \textbf{L} & 3 & 5 & 7 & 9 & 3 & 5 & 7 & 9 & 3 & 5 & 7 & 9\\
\hline
   Markov+DBS & 0.611 & 0.585 & 0.553 & 0.524 & 1.000 & 0.425 & 0.190 & 0.103 & 0.805 & 0.505 & 0.372 & 0.313 \\
\hline
   NASR+DBS & \textbf{0.624} & \textbf{0.604} & \textbf{0.565} & \textbf{0.528} & 1.000 & 0.422 & 0.248 & 0.173 & \textbf{0.812} & 0.513 & 0.406 & 0.350 \\
\hline
   \methodname-LSTM & 0.621 & 0.585 & 0.545 & 0.510 & 1.000 & \textbf{0.741} & \textbf{0.621} & \textbf{0.508} & 0.810 & \textbf{0.663} & \textbf{0.583} & \textbf{0.509} \\
\hline
   \methodname-Samp & 0.622 & 0.582 & 0.537 & 0.507 & 1.000 & 0.704 & 0.590 & 0.467 & 0.811 & 0.643 & 0.564 & 0.487\\ 
\hline
\end{tabular}
       }
\caption{Epcot Dataset}
\label{table_epcot_L}
\end{subtable}
\caption{Comparison With Varying Length of Recommended Itineraries}
\label{tab:comp_L}
\end{table*}
\subsection{Hyperparameter Tuning}
We now describe the hyperparameter values used in \textit{\methodname-LSTM} and \textit{\methodname-Samp}.Recall that our algorithm first obtains POI graph embeddings through two separate graph autoencoders. It then trains two LSTM models. Finally separate itineraries are generated through $k$ different prominent POIs, through the use of either an LSTM based technique (\textit{\methodname-LSTM}) or a sampling based technique (\textit{\methodname-Samp}).

\textbf{Graph Autoencoders and ITRNet:} For the graph autoencoder, the embedding dimensions $Z_C$ and $Z_D$ were 12 and 24, respectively. The autoencoders were trained with learning rate of 0.05 and 0.01 respectively, using the Adam optimizer. The hidden layer size for both the forward and backward LSTM models was 32. The dimension of the MLP layer was 30. Here, the learning rate was set 0.001 for the whole model, using the Adam optimizer as before. Both the LSTM models were trained with a batch size of 32.

\textbf{\methodname-LSTM:} To generate fixed length itineraries containing $L$ POIs, we first generate the half itinerary as prescribed in the algorithm setting $L_{max}$ to $L-1$, and then the other half itinerary is generated such that the length of the total itinerary is $L$. Also $L_{max}$ is set to the length of the longest itinerary found in the training dataset.

\textbf{\methodname-Samp:} We start with an initial itinerary consisting of $L$ POIs by placing the intermediate prominent POI at a random position between $2$ to $L-1$. We ignore the insert and delete operations here as those operations would change the length. \textit{Replacement} or \textit{Swap and Replace} operations are performed each with probability 0.5. For prominent POI, we do not use replace operation and only apply \textit{Swap and Replace} operation with probability 0.5. The sampling algorithm is run for $5(L-2)$ iterations in total. This ensures that when the search space is larger (i.e., larger $L$), the algorithm runs more iterations to achieve good quality.
\subsection{Performance Comparison}
We now discuss performance in terms of the evaluation metrics considered. We first consider the average popularity and diversity of the recommended itineraries independently, after that we consider the combined score to assess the performance of the system to return multiple alternative itineraries. Next we evaluate the effect on performance if we vary the length of the recommended itineraries and also if we vary the no. of itineraries to be recommended. We also compare the running times of the variations of \methodname~ and also compare them with the baselines.

\subsubsection{Considering Popularity and Diversity Independently}
Table \ref{tab:table_f1_pf1_div_all} shows the popularity (using both F1 score and pairs-F1 score) and the diversity of the  itineraries recommended by each approach. We see that the average F1 and pairs-F1 scores of both of our approaches are similar to those of the competitors which are the state-of-the-art for returning most popular itineraries. On the other hand, the average diversity of the recommended itineraries provided by our approaches far exceed those of the competitors in all datasets. For example in the Edinburgh dataset, \textit{\methodname-LSTM} and \textit{\methodname-Samp} provide 19.13\% and 12.60\% higher average diversity, respectively, than the nearest competing baseline. This shows that our approaches provide much more diverse itineraries while keeping the popularity of the recommended itineraries on par with the other baselines. In other words, these baselines primarily focus on popularity, and the competitive F1 and pairs-F1 scores show that our approaches generate diverse itineraries without compromising on the popularity of the recommended itineraries, thus providing quality alternative itineraries.
\begin{table}[ht]
\begin{subtable}[t]{\columnwidth}
\resizebox{\columnwidth}{!}{
\begin{tabular}[t]{|c|c|c|c|c|c|}
\hline
  & \multicolumn{5}{|c|}{$\boldsymbol{\alpha}$}\\
\hline
   & 0.1 & 0.3 & 0.5 & 0.7 & 0.9\\
\hline
  Markov+DBS & 0.540 & 0.552 & 0.563 & 0.575 & 0.586\\
\hline
  NASR+DBS & 0.637 & 0.624 & 0.610 & 0.596 & 0.583\\
\hline
  \methodname-LSTM & \textbf{0.747} & \textbf{0.710} & \textbf{0.673} & \textbf{0.636} & \textbf{0.598}\\
\hline
  \methodname-Samp & 0.710 & 0.680 & 0.651 & 0.621 & 0.592\\
\hline
\end{tabular}
      }
\caption{Edinburgh Dataset}
\end{subtable}

\bigskip 

\begin{subtable}[t]{\columnwidth}
\resizebox{\columnwidth}{!}{
\begin{tabular}[t]{|c|c|c|c|c|c|}
\hline
  & \multicolumn{5}{|c|}{$\boldsymbol{\alpha}$}\\
\hline
  & 0.1 & 0.3 & 0.5 & 0.7 & 0.9\\
\hline
  Markov+DBS & 0.441 & 0.473 & 0.505 & 0.537 & 0.569\\
\hline
  NASR+DBS & 0.440 & 0.477 & 0.513 & 0.549 & 0.586\\
\hline
  \methodname-LSTM & \textbf{0.726} & \textbf{0.694} & \textbf{0.663} & \textbf{0.632} & \textbf{0.601}\\
\hline
  \methodname-Samp & 0.692 & 0.668 & 0.643 & 0.619 & 0.594\\
\hline
\end{tabular}
      }
\caption{Epcot Dataset}
\end{subtable}
\caption{Comparison with Combined Metric}
\label{tab:combined_metric}
\end{table}

\begin{table*}
\begin{subtable}[t]{2\columnwidth}
\resizebox{\columnwidth}{!}{
\begin{tabular}[t]{|c|c|c|c|c|c|c|c|c|c|c|c|c|c|c|c|}
\hline
   & \multicolumn{5}{|c|}{\textbf{F1 Score}} & \multicolumn{5}{|c|}{\textbf{Diversity}} & \multicolumn{5}{|c|}{\textbf{Comb ($\boldsymbol{\alpha}$ = 0.5)}}\\
\hline
   \textbf{k} & 1 & 2 & 3 & 4 & 5 & 1 & 2 & 3 & 4 & 5 & 1 & 2 & 3 & 4 & 5\\
\hline
   Markov+DBS & 0.600 & 0.589 & \textbf{0.592} & \textbf{0.585} & \textbf{0.586} & -- & 0.564 & 0.534 & 0.487 & 0.459 & -- & 0.576 & 0.563 & 0.536 & 0.522\\
\hline
   NASR+DBS & 0.599 & 0.585 & 0.576 & 0.573 & 0.574 & -- & 0.626 & 0.643 & 0.644 & 0.639 & -- & 0.605 & 0.610 & 0.608 & 0.606\\
\hline
   \methodname-LSTM & \textbf{0.608} & \textbf{0.589} & 0.580 & 0.572 & 0.568 & -- & \textbf{0.764} & \textbf{0.766} & \textbf{0.774} & \textbf{0.787} & -- & \textbf{0.676} & \textbf{0.673} & \textbf{0.673} & \textbf{0.677}\\
\hline
   \methodname-Samp & 0.587 & 0.579 & 0.577 & 0.571 & 0.568 & -- & 0.709 & 0.724 & 0.737 & 0.745 & -- & 0.644 & 0.651 & 0.654 & 0.657\\
\hline
\end{tabular}
       }
\label{table_edin_K}
\caption{Edinburgh Dataset}
\end{subtable}


\begin{subtable}[t]{2\columnwidth}
\resizebox{\columnwidth}{!}{
\begin{tabular}[t]{|c|c|c|c|c|c|c|c|c|c|c|c|c|c|c|c|}
\hline
    & \multicolumn{5}{|c|}{\textbf{F1 Score}} & \multicolumn{5}{|c|}{\textbf{Diversity}} & \multicolumn{5}{|c|}{\textbf{Comb ($\boldsymbol{\alpha}$ = 0.5)}}\\
\hline
   \textbf{k} & 1 & 2 & 3 & 4 & 5 & 1 & 2 & 3 & 4 & 5 & 1 & 2 & 3 & 4 & 5\\
\hline
   Markov+DBS & 0.593 & 0.584 & 0.585 & 0.582 & 0.583 & -- & 0.465 & 0.425 & 0.378 & 0.366 & -- & 0.524 & 0.505 & 0.480 & 0.475\\
\hline
   NASR+DBS & \textbf{0.612} & \textbf{0.607} & \textbf{0.604} & \textbf{0.604} & \textbf{0.602} & -- & 0.475 & 0.422 & 0.425 & 0.431 & -- & 0.541 & 0.513 & 0.515 & 0.516\\
\hline
   \methodname-LSTM & 0.594 & 0.591 & 0.585 & 0.581 & 0.575 & -- & \textbf{0.714} & \textbf{0.741} & \textbf{0.754} & \textbf{0.765} & -- & \textbf{0.652} & \textbf{0.663} & \textbf{0.668} & \textbf{0.670}\\
\hline
   \methodname-Samp & 0.591 & 0.592 & 0.582 & 0.573 & 0.570 & -- & 0.688 & 0.704 & 0.733 & 0.737 & -- & 0.640 & 0.643 & 0.653 & 0.653\\
\hline
\end{tabular}
       }
\label{table_epcot_K}
\caption{Epcot Dataset}
\end{subtable}
\caption{Effect of Varying The No. of Itineraries Generated (\textit{k})}
\label{tab:comp_k}
\end{table*}
\subsubsection{Considering Combined Popularity and Diversity Score}
We vary $\alpha$ in the combined metric from $0.1$ to $0.9$ and show the results for each value in Table~\ref{tab:combined_metric}. Due to space constraints, we only present the results for two datasets, one from each group. Results on the other datasets show similar trends. Again the length of the recommended itineraries is set to 3 and the no. of itineraries recommended is set to 5. The scores of the combined metric is shown in Table \ref{tab:combined_metric}.

We observe that both variants of \textit{\methodname} outperform the competitors even when $\alpha$ is set to 0.9, i.e., the popularity is given a much higher importance than diversity. When both are given equal importance i.e., $\alpha = 0.5$, we see that in the Epcot dataset \textit{\methodname-LSTM} and \textit{\methodname-Samp} outperform the nearest competing baseline by 29.24\% and 25.34\%, respectively. 



\subsubsection{Effect of Varying Length of Recommended Itineraries}
We vary the length $L$ of the recommended itineraries as 3, 5, 7, and 9. Table \ref{tab:comp_L} shows the results. We show the results for the Edinburgh and Epcot datasets (other datasets also follow similar trend).
Note that, for all values of $L$, the average F1 and pairs-F1 score remain similar to those of the competitors that primarily focus on providing popular itineraries. However, the diversity of the recommended itineraries by \textit{\methodname-LSTM} and \textit{\methodname-Samp} are significantly higher than these competitors.


We observe that, as the value of $L$ increases, both \textit{\methodname-LSTM} and \textit{\methodname-Samp} outperform the nearest competing baseline by a greater margin. For example, for the Edinburgh dataset, \textit{\methodname-LSTM} provides 19.13\% increase in diversity and 10.33\% increase in the combined score for $L = 5$, whereas it provides a 136\% increase  in diversity and a 44.38\% increase in the combined score for $L = 9$. Similarly, for \textit{\methodname-Samp} in the Edinburgh dataset, we see a 12.60\% increase in diversity and 6.72\% increase in the combined score for $L = 5$, whereas a 107\% increase in diversity and 34.55\% increase in the combined score for $L = 9$. This is primarily because  the average diversity provided by the baselines significantly drops for larger $L$. Note that, for the case when itinerary length is $3$ (including $s$ and $d$), diversity for each approach is maximum (i.e., $1$) which is because the only intermediate POI in each itinerary is different from the other recommended itineraries.



\subsubsection{Effect of k} 

Here we set the length of itineraries recommended, $L$ to 5. Again we show the results in two datasets for space constraints, taking one each from the two different domains. Other datasets show similar trends. The results are shown in Table \ref{tab:comp_k}.

Our proposed approaches consistently achieve higher diversity even for larger $k$.   The average F1 score of the recommended itineraries slightly drop for all approaches, however, our approaches are comparable to the baselines. Consequently, we see that both \textit{\methodname-LSTM} and \textit{\methodname-Samp} provide significantly higher combined scores. For example, in the Edinburgh dataset, \textit{\methodname-LSTM} provides 11.74\% , 10.32\%, 10.69\% and 11.72\% higher combined score with $\alpha = 0.5$ for $k = 2, 3, 4$ and $5$, respectively. Also \textit{\methodname-Samp} provides 9.53\%, 6.72\%, 7.57\% and 8.42\% higher combined scores for $k=2, 3, 4$ and $5$, respectively, in the same dataset.

\subsection{Running Time Comparison}

We run all the algorithms on the same machine equipped with Intel core-i7 8565U CPU, 16GB RAM. We record the average time per query (in seconds) for five folds of the dataset, and report the average time per query taken across the five folds.

We vary the length of recommended itineraries, $L$ as 3,5,7 and 9 and keep the no of recommended itineraries $k$ as 3. As the Melbourne dataset has the maximum no. of POIs and Disneyland dataset has maximum no. trips, we show the results for these two datasets to depict the scalability of the algorithms. The results are shown in Table \ref{tab:execution_time}.

\begin{table}[ht]
\begin{subtable}[t]{\columnwidth}
\resizebox{\columnwidth}{!}{
\begin{tabular}[t]{|c|c|c|c|c|}
\hline
  & \multicolumn{4}{|c|}{\textbf{Execution Time (Seconds)}}\\
\hline
  \textbf{L} & 3 & 5 & 7 & 9\\
\hline
  Markov+DBS & 0.700 & 1.062 & 1.424 & 1.860\\
\hline
  NASR+DBS & 0.060 & 0.167 & 0.280 & 0.413\\
\hline
  \methodname-LSTM & 0.093 & 0.190 & 0.293 & 0.411\\
\hline
  \methodname-Samp & 0.084 & 0.491 & 0.868 & 1.257\\
\hline
\end{tabular}
      }
\caption{Melbourne Dataset}
\end{subtable}

\bigskip 

\begin{subtable}[t]{\columnwidth}
\resizebox{\columnwidth}{!}{
\begin{tabular}[t]{|c|c|c|c|c|}
\hline
  & \multicolumn{4}{|c|}{\textbf{Execution Time (Seconds)}}\\
\hline
  \textbf{L} & 3 & 5 & 7 & 9\\
\hline
  Markov+DBS & 0.152 & 0.171 & 0.204 & 0.245\\
\hline
  NASR+DBS & 0.057 & 0.166 & 0.286 & 0.429\\
\hline
  \methodname-LSTM & 0.088 & 0.199 & 0.303 & 0.416\\
\hline
  \methodname-Samp & 0.074 & 0.474 & 0.827 & 1.173\\
\hline
\end{tabular}
      }
\caption{Disneyland Dataset}
\end{subtable}
\caption{Query Execution Time (Seconds)}
\label{tab:execution_time}
\end{table}
We observe that \textit{\methodname-LSTM} and \textit{NASR+DBS} have similar execution times in both datasets; whereas  \textit{\methodname-Samp} takes more time than the other approaches. We also see that the execution time increases with the increase of $L$. However as trip recommendation systems generally do not recommend excessively long routes to users \cite{chen2011discovering}, this increasing trend of query execution time is quite acceptable. We observe that although the Melbourne dataset have almost three times more POIs than the Disneyland dataset, the average execution time per query remains similar for \textit{NASR+DBS, \methodname-LSTM} and \textit{\methodname-Samp}. Hence, the running times of these three algorithms are not significantly  influenced by the no. of POIs; whereas the execution time of \textit{Markov+DBS}  increases substantially as the no. of POIs increases.

\label{sub:experiments}
\section{Conclusion}
This paper proposed two deep-learning-based approaches that learns to recommend top-$k$ alternative itineraries for a given source and destination. We first developed \textit{Itinerary Net} (ITRNet) that estimates the likelihood of POIs on an itinerary by using graph autoencoders and two (forward and backward) LSTMs. 
Based on the ITRNet, we have developed two variants, \textit{\methodname-LSTM} and \textit{\methodname-Samp}, to recommend $k$ alternative itineraries using historical trips without requiring any explicit popularity or diversity modeling. Our \textit{\methodname-Samp} solution can also trivially incorporate various user defined constraints. Extensive experiments using real-world datasets show that the \textit{\methodname-LSTM} and \textit{\methodname-Samp} outperform the best performing baselines by up to 29.24\% and 25.34\%, respectively, for the default settings w.r.t. the combined popularity and diversity measure. In the future, we plan to incorporate user personalization leveraging information of the historical routes visited by the querying user.\label{sub:conclusion}
\bibliographystyle{IEEEtran}
\bibliography{refs}

%







\begin{IEEEbiography}[{\includegraphics[width=1.1 in,height=1.1 in,clip,keepaspectratio]{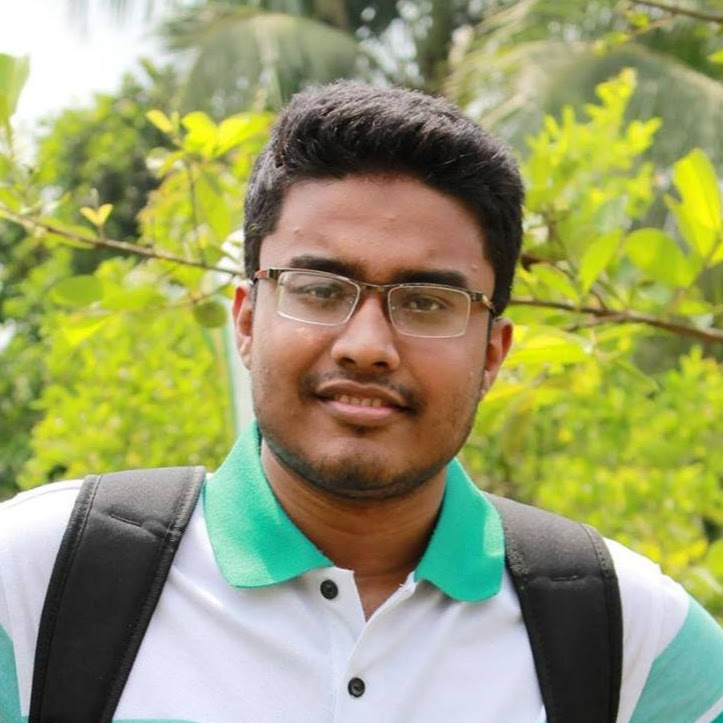}}]{Syed Md. Mukit Rashid} is a Lecturer at Bangladesh University of Engineering and Technology (BUET), Dhaka since October 2019. He received his B.Sc. degree in CSE from BUET in April 2019. He is currently pursuing his M.Sc. degree in CSE from the same university. His research interests are computer vision, applied machine learning and network security.
\end{IEEEbiography}

\begin{IEEEbiography}[{\includegraphics[width=1.2in,height=1.2in,clip,keepaspectratio]{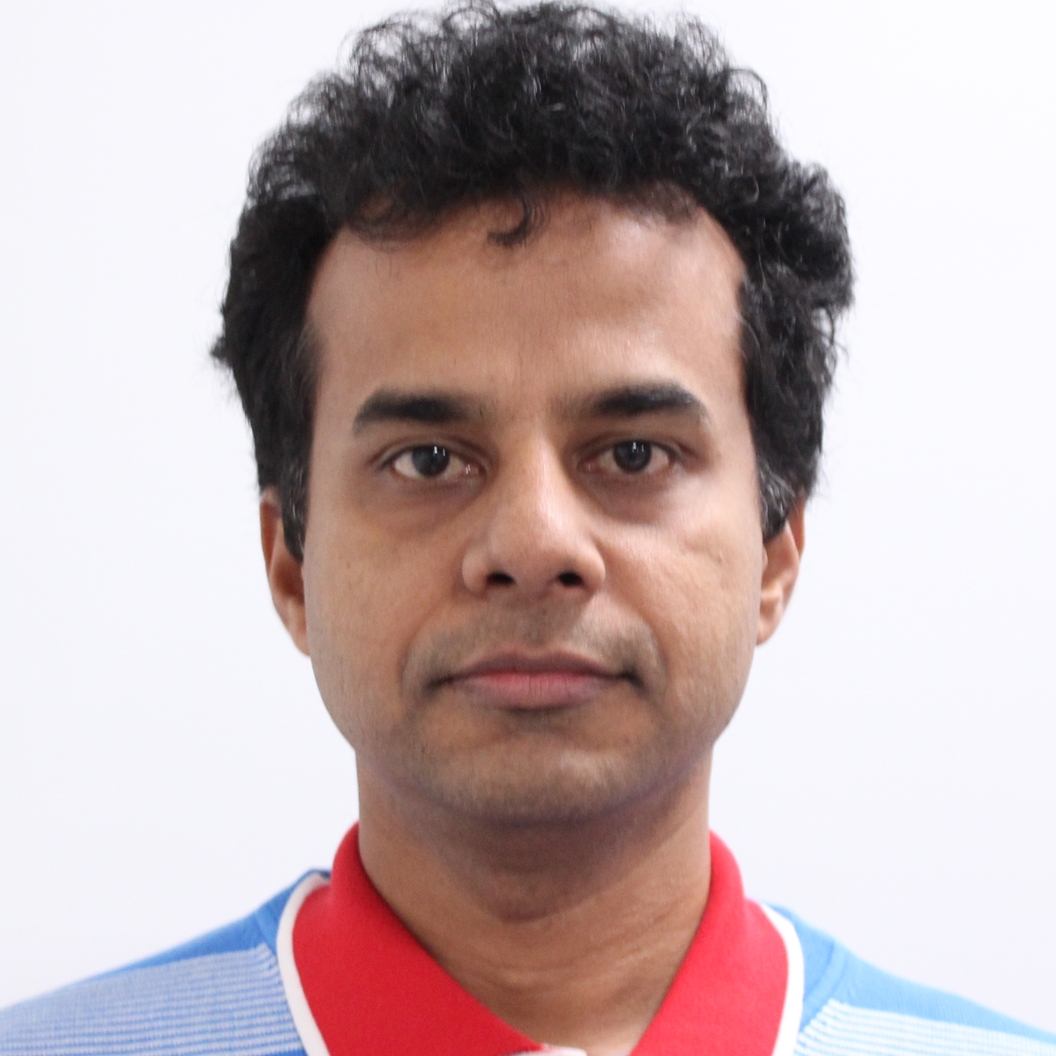}}]{Mohammed Eunus Ali} is a Professor at Bangladesh University of Engineering and Technology (BUET), Dhaka since May 2014. He is the group leader of Data Science and Engineering Research Lab (datalab.buet.io) in the department of Computer Science and
Engineering at BUET. He received his PhD from the University of Melbourne in 2010. His research falls in the intersection of data management and machine learning. His research areas cover a wide range of topics in database systems and information management that include spatial databases, practical machine learning, social media analytics, and mobile health. His research has been published in top ranking journals and conferences such as the VLDB Journal, PVLDB, ICDE, CIKM, EDBT, SIGCHI, and UbiComp.
\end{IEEEbiography}

\begin{IEEEbiography}[{\includegraphics[width=1.2in,height=1.2in,clip,keepaspectratio]{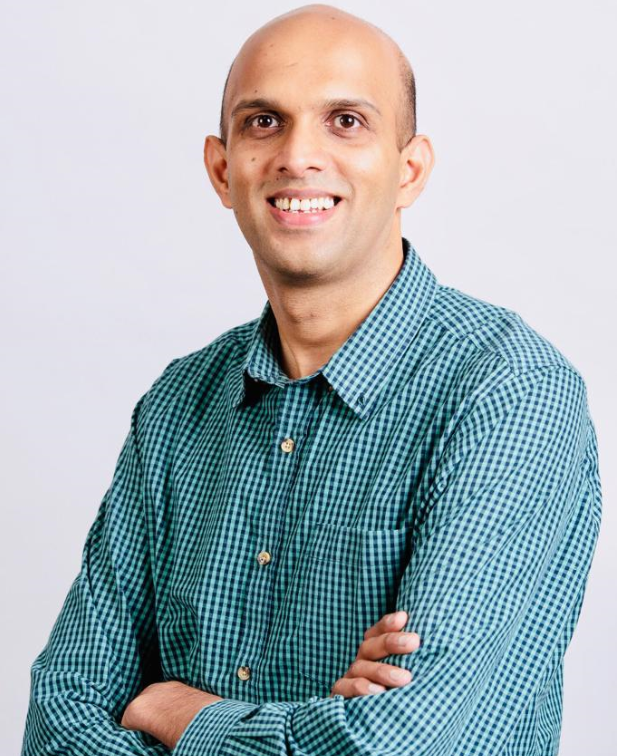}}]{Muhammad Aamir Cheema} is an ARC Future Fellow, an Associate Professor at the Faculty of Information Technology,  Monash University, Australia. He obtained his PhD from UNSW Australia in 2011. He is the recipient of 2012 Malcolm Chaikin Prize for Research Excellence in Engineering, 2013 Discovery Early Career Researcher Award, 2014 Dean’s Award for Excellence in Research by an Early Career Researcher, 2018 Future Fellowship, 2018 Monash Student Association Teaching Award and 2019 Young Tall Poppy Science Award. He has also won two CiSRA best research paper of the year awards, two invited papers in the special issue of IEEE TKDE on the best papers of ICDE, and three best paper awards at ICAPS 2020, WISE 2013 and ADC 2010, respectively. 
\end{IEEEbiography}

\end{document}